\def\year{2021}\relax
\documentclass[letterpaper]{article} 
\usepackage{aaai21}  
\usepackage{times}  
\usepackage{helvet} 
\usepackage{courier}  
\usepackage[hyphens]{url}  
\usepackage{graphicx} 
\usepackage{natbib}  
\usepackage{caption} 
\usepackage{epsfig}
\usepackage{graphicx}
\usepackage{amsmath}
\usepackage{amssymb}
 \usepackage{multirow}
 \usepackage{enumitem}
 \usepackage{caption,subcaption}
\usepackage{booktabs} 
\usepackage[font=small,labelfont=bf]{caption}
\usepackage{wrapfig}

\usepackage[pagebackref=true,breaklinks=true,letterpaper=true,colorlinks,bookmarks=false]{hyperref}
\usepackage{subfiles}

\newcommand\numberthis{\addtocounter{equation}{1}\tag{\theequation}}

\renewcommand\cite{\citet}

\newcommand{\HB}[1]{}
\newcommand{\Sam}[1]{}
\newcommand{\Anirudh}[1]{}
\newcommand{\Hugo}[1]{}
\newcommand{\Animesh}[1]{}
\newcommand{\Florian}[1]{}

\newcommand{\algoName}{DIBS } 

\title{Diversity inducing Information Bottleneck in Model Ensembles }
\author{
      Samarth Sinha$^*$\textsuperscript{\rm 1}
    Homanga Bharadhwaj$^*$ \textsuperscript{\rm 1},
    Anirudh Goyal \textsuperscript{\rm 2},
    Hugo Larochelle \textsuperscript{\rm 2,3},
       Animesh Garg\textsuperscript{\rm 1},
       Florian Shkurti  \textsuperscript{\rm 1}
}
\affiliations{

   \textsuperscript{\rm 1} University of Toronto, Vector Institute\\
   \textsuperscript{\rm 2} Mila, University of Montreal \\
   \textsuperscript{\rm 3} Google Brain

}

\begin{document}

\maketitle

\begin{abstract}
Although deep learning models have achieved state-of-the art performance on a number of vision tasks, generalization over high dimensional multi-modal data, and reliable predictive uncertainty estimation are still active areas of research.  Bayesian approaches including Bayesian Neural Nets (BNNs) do not scale well to modern computer vision tasks, as they are difficult to train, and have poor generalization under dataset-shift~\cite{lakshminarayanan2017simple,ensemblesbest}. This motivates the need for effective ensembles which can generalize and give reliable uncertainty estimates. In this paper, we target the problem of generating effective ensembles of neural networks by encouraging diversity in prediction. We explicitly optimize a diversity inducing adversarial loss for learning the stochastic latent variables and thereby obtain diversity in the output predictions necessary for modeling multi-modal data. 
We evaluate our method on benchmark datasets: MNIST, CIFAR100, TinyImageNet and MIT Places 2, and compared to the most competitive baselines show significant improvements in classification accuracy, under a shift in the data distribution and in out-of-distribution detection.
: over $10\%$ relative improvement in classification accuracy, over $5\%$ relative improvement in generalizing under dataset shift, and over $5\%$ better predictive uncertainty estimation as inferred by efficient out-of-distribution (OOD) detection.
\end{abstract}
\section{Introduction}

Deep Neural Networks (DNNs) have achieved state-of-the-art performance in a wide variety of vision tasks, where the goal is to perform a single task efficiently \cite{resnet,wideresnet,drn,maskrcnn}. However, most state-of-the-art approaches in computer vision, train a single network for solving a particular task, which may not generalize when there is a change in the input distribution during evaluation. Related to the issue of generalization, the notion of predictive uncertainty quantification remains an open problem. To achieve this, it is important for the learned model to be uncertainty-aware, or to \textit{know what it does not know}. One of the ways of estimating this is to show the network out-of-distribution (OOD) examples, and evaluate it on the effectiveness of the model to determine OOD samples~\cite{ood}.

Bayesian Neural Networks (BNNs)~\cite{bnn} and MC-dropout~\cite{mcdropout} are theoretically motivated Bayesian methods, and have seen many applications in modeling predictive uncertainty. However, BNNs are: difficult to train, do not scale well to high-dimensional data, and do not perform well under dataset-shift~\cite{lakshminarayanan2017simple,anonymouslosslandscape,vaal}.
In addition, the choice of priors over the model weights is a crucial factor in their effectiveness. \cite{ncp,lakshminarayanan2017simple}. MC-dropout is a fast and easy to train alternative to BNNs, and can be interpreted as an ensemble model followed by model averaging.
However, recent works highlight its limitations in deep learning for uncertainty prediction \cite{vaal,sener2017active}, generalization, and predictive accuracy~\cite{aleatoricvision,lakshminarayanan2017simple}.

Our work is motivated to provide better generalization and provide reliable uncertainty estimates, which we obtain from inferring multiple plausible hypotheses that are \textit{sufficiently} diverse from each other.
This is even more important in cases of high dimensional inputs, like images, because the data distribution is inherently multimodal. 
Ensemble learning is a natural candidate for learning multiple hypotheses from data.  We address the problem of introducing diversity among the different ensemble components~\cite{diverseensembles} and at the same time ensuring that the predictions balance data likelihood and diversity. 
To achieve this, we propose an adversarial diversity inducing objective with a information bottleneck (IB) constraint~\cite{tishby2000information,vib} to enforce forgetting the input as much as possible, while being predictive about the output. IB~\cite{tishby2000information} formalizes this in terms of minimizing the mutual information (MI) between the bottleneck representation layer with the input, while maximizing its MI
with the correct output, 
which has been shown to improve generalization in neural networks~\cite{vib,achille2018information,goyal2019infobot}.

Recent methods in ensemble learning~\cite{lakshminarayanan2017simple,anonymouslosslandscape} illustrate the drawbacks of applying classical ensembling techniques like bootstrapping~\cite{bagging} to deep neural nets. 
A recent paper~\cite{anonymouslosslandscape} analyzes the empirical success of ensembling using random initializations compared to Bayesian uncertainty estimation techniques such as BNNs~\cite{bnn}, and MC-dropout~\cite{mcdropout} and arrives at the conclusion that random ensembles sucessfully identify different modes in the data but they do not fit accurately to any mode while Bayesian methods fit accurately but to just one mode in the data.
This motivates the need for an ensembling approach that both identifies different modes and fits accurately to each mode, thereby achieving high accuracy, high generalization, and precise uncertainty estimates.

\begin{figure}[t]
    \centering
    \includegraphics[width=0.9\columnwidth]{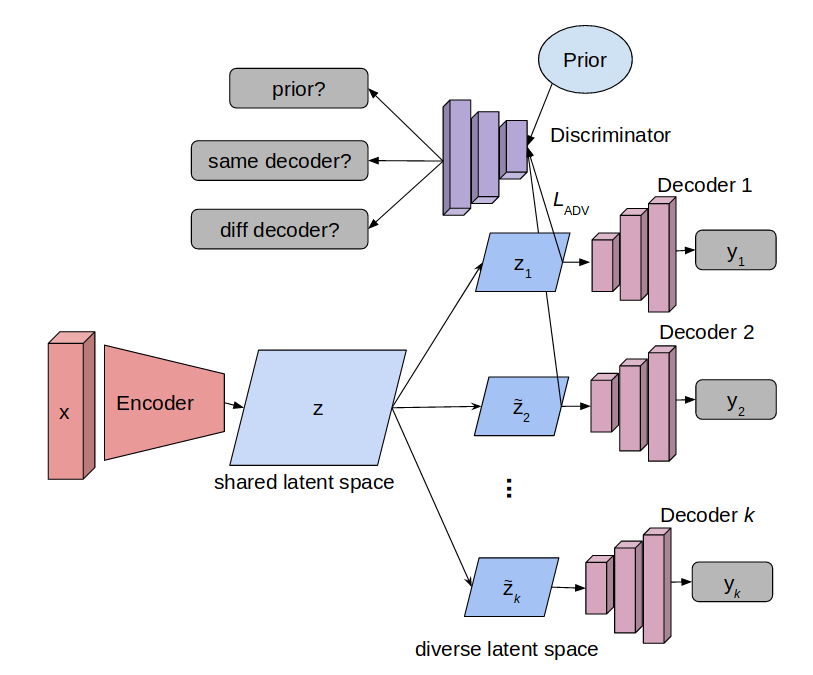}
    \caption{The basic structure of our proposed diverse ensembles approach. The input $X$ is mapped to a shared latent variable $Z$ through a deterministic encoder. The shared $Z$ is mapped to $K$ different stochastic variables $\tilde{Z}_i$ which finally map to the $K$ different outputs $Y_i$, $\forall i\in[1,..K]$}
    \label{fig:main}
\end{figure}

We propose a principled scheme of ensemble learning, by jointly maximizing data likelihood, constraining information flow through a bottleneck to ensure the ensembles capture only $relevant$ statistics of the input, and maximizing a diversity inducing objective to ensure that the multiple plausible hypotheses learned are diverse. 
Instead of $K$ different neural nets, we have $K$ different 
stochastic decoder heads, as shown in Fig.~\ref{fig:main}. 
We explicitly maximize diversity among the ensembles by an adversarial loss. 
Our ensemble learning scheme has several advantages as compared to randomly initialized ensembles and Bayesian NNs 
 since the joint encoder helps us in learning shared `basic' representations that will be useful for all the decoders. 
 We are able to explicitly control the flow of information from the encoder to each of the decoders
 during training.
 Most importantly, we can explicitly enforce diversity among the decoder heads and do not have to rely on 
 random initialization or a prior on the weights to yield diverse output. 
 We show that this diversity enforcing objective helps capture the multiple modes in the underlying data distribution.


In summary, we claim the following contributions:
\begin{enumerate}[
    topsep=0pt,
    noitemsep,
    leftmargin=*,
    itemindent=12pt]
\item We introduce diversity among the ensemble members through a novel adversarial loss that encourages samples from different stochastic latent variables to be separated and samples from the same stochastic latent variable to be close to each other. 
\item We generalize the VIB~\cite{vib} formulation to multiple stochastic latent variables and balance diversity with high likelihood by enforcing an information botleneck between the stoachastic latent variables, $\tilde{Z}_i$, and the input $X$.
\end{enumerate}

Through extensive experimentaton, we demonstrate better generalization to dataset shift, better performance when training with few labels as compared to state-of-the-art baselines, 
and show better uncertainty estimation on OOD detection. 
Finally we demonstrate that we achieve consistently better performance with respect to baselines when varying the number of decoders ($K$) in the ensemble.

\section{Preliminaries}


\subsection{Mutual Information, Information Bottleneck}
Mutual Information (MI) is a measure of dependence between two random variables. The MI between random variables $X$ and $Y$ is defined as the KL divergence between the joint distribution and the product of the marginals:
\begin{align*}
    \mathcal{I}(X,Y) 
    &= KL(\mathbb{P}_{XY}|| \mathbb{P}_X \mathbb{P}_Y)  \numberthis \label{eqn:MIdef} 
\end{align*}
By the definition of KL divergence between two probability distributions $\mathbb{P}$ and $\mathbb{Q}$, $ KL(\mathbb{P}||\mathbb{Q}) = \mathbb{E}_{\mathbb{P}}[\log d\mathbb{P}/d\mathbb{Q}]$, we have:
\begin{align*}
    \mathcal{I}(X,Y) 
    &= \int\int  p(x,y)\log\frac{p(x,y)}{p(x)p(y)} dx dy \numberthis \label{eqn:MIdef1}
\end{align*}
In the variational information bottleneck (VIB) literature~\cite{vib}, given input-output pairs $X$ and $Y$, the goal is to learn a compressed latent space $Z$ to maximize the mutual information between $Z$ and $Y$, while minimizing the mutual information between $Z$ and $X$ to learn a representations $Z$ that sufficiently forgets the spurious correlations that may exist in the input $X$, while still being predictive of $Y$. More formally: 

\begin{align*}
    \max_{\theta} \; \mathcal{I}(Z,Y;\theta) \quad \text{s.t.}  \quad \mathcal{I}(X,Z;\theta)\leq I_c
\end{align*}

Here $I_c$ is some information constraint. This constrained optimization can be solved through Lagrange multipliers. 

\subsection{Notation}


In this paper, we consider a network with a shared encoder $f(\cdot)$ and multiple stochastic task-specific decoders $g_i(\cdot)$ where both $f(\cdot)$ and $g_i(\cdot)$ are parameterized using neural networks.
Each encoder encodes an image, $X$, to a shared latent space, $Z$, which is then used by each task specific decoder to obtain a prediction $Y_i$, where $Y_i$ is the $i$-th prediction from the model. Fig \ref{fig:main} describes the architecture visually.


\section{\algoName: Diverse Information Bottleneck in Ensembles}

We propose a method for ensemble learning~\cite{ensemblesold1} by promoting diversity among pairwise latent ensemble variables and by enforcing an information bottleneck~\cite{vib} between each latent $\tilde{Z}_i$ and the input $X$. Formally, we consider a set of $K$ decoders $\{\theta_1,...\theta_K\}\sim p(\Theta)$ sampled from some given initial distribution $p(\Theta)$. Given input data $X$, we want to learn a shared encoding $Z$, and $K$ decoders that map the latent state $Z$ to $\tilde{Z}_i$'s and  $\tilde{Z}_i$'s to the $K$ output predictions $\{Y_1,...,Y_K\}$.  

We posit that for effective learning through ensembles, there must be some diversity among the members of the ensemble, since each ensemble member is by assumption a weak learner, and individual performance is not as important as collective performance~\cite{diverseensembles}. 
However, promoting diversity randomly among the members is likely to result in uninformative/irrelevant aspects of data being captured by them.
Hence, in addition to task-specific standard likelihood maximization, we introduce the need for a diversity enforcing constraint, and a bottleneck constraint. To accomplish the latter, we build upon the Variational Information Bottleneck (VIB) formulation~\cite{vib} by constraining the information flow from input $X$ to the outputs $Y_i$, we introduce the information bottleneck term $-\mathcal{I}(\tilde{Z}_i,X;\theta)$. For diversity maximization between ensembles, we design an anti-clustering and diversity-inducing generative adversarial loss, described in the next section.

\subsection{Adversarial Model for Diversity Maximization}
\label{sec:diversityloss}

We adopt an adversarial learning approach based on the intuition of diversity maximization among the $K$ models. Our method is inspired by Adversarial Autoencoders~\cite{adversarialvae}, which proposes a natural scheme for combining adversarial training with variational inference. Here, our aim is to maximize separation in distribution between ensemble pairs $q(\tilde{z}_i|x),q(\tilde{z}_j|x)$, such that samples $(\hat{z}_1,\hat{z}_2)$ are indistinguishable to a discriminator if $\hat{z}_1\sim q(\tilde{z}_i|x)$, $\hat{z}_2\sim q(\tilde{z}_j|x)$  with $i=j$ and they are distinguishable if $i\neq j$. To this end, we frame the adversarial loss, such that the $K$ generators $q(\tilde{z}_i|x) \; \forall i\in [1,..,K]$ trick the discriminator into thinking that samples from $q(\tilde{z}_i|x)$ and $q(\tilde{z}_j|x)$ are samples from different distributions.


We start with $r(\tilde{z})$, a prior distribution on $\tilde{z}$. In our case, this is normal $\mathcal{N}(0, \mathbb{I})$, but more complex priors are also supported, in the form of implicit models. We want to make each encoder $q(\tilde{z}_i|x)$ to be close in distribution to this prior, but sufficiently far from other encoders, so that overlap is minimized. Unlike typical GANs~\cite{gan}, the discriminator of our diversity inducing loss takes in a pair of samples ($\hat{z}_1,\hat{z}_2$) instead of just one sample. Hence, we have the following possibilities for the different sources of a pair of latents: 1) $\hat{z}_1\sim r(\tilde{z})$ and $\hat{z}_2\sim r(\tilde{z})$, 2) $\hat{z}_1\sim q(\tilde{z}_i|x)$ and $\hat{z}_2\sim r(\tilde{z})$, 3)  $\hat{z}_1\sim q(\tilde{z}_i|x)$ and $\hat{z}_2\sim q(\tilde{z}_i|x)$, and 4) $\hat{z}_1\sim q(\tilde{z}_i|x)$ and $\hat{z}_2\sim q(\tilde{z}_j|x)$, with $i \neq j$ 


Let $D(\cdot)$ denote the discriminator, which is a feed-forward neural network that takes in a pair $(\hat{z}_1,\hat{z}_2)$ as input and outputs a $0$ (fake) or a $1$ (real). There are $K$ generators corresponding to each $q(\tilde{z}_j|z)$, and the deterministic encoder $z = f(x)$. We denote the parameters of all these generators, as well as the deterministic encoder as $G$, to simplify notation. These generators are trained by minimizing the following loss over $G$:
\begin{align*}
    L_G &= 
    \mathbb{E}_{\hat{z}_1 \sim q(\tilde{z}_i|x), \; \hat{z}_2\sim q(\tilde{z}_j|x)}[\log D(\hat{z}_1,\hat{z}_2)]\\
    &+ \mathbb{E}_{\hat{z}_1 \sim r(\tilde{z}), \; \hat{z}_2\sim q(\tilde{z}_i|x)}[\log (1- D(\hat{z}_1,\hat{z}_2))]\\
    &+ \mathbb{E}_{\hat{z}_1 \sim q(\tilde{z}_i |x), \; \hat{z}_2\sim q(\tilde{z}_i |x)}[\log (1 -D(\hat{z}_1,\hat{z}_2))]
    \numberthis 
    \label{eqn:diversitygen}
\end{align*}
\noindent Given a fixed discriminator $D$, the first term encourages pairs of different encoder heads to be distinguishable. The second term encourages each encoder to overlap with the prior. The third term encourages samples from the same encoder to be indistinguishable. 

On the other hand, given a fixed generator $G$, the discriminator is trained by maximizing the following objective function with respect to $D$:
\begin{align*}
    L_D &= \mathbb{E}_{\hat{z}_1 \sim r(\tilde{z}), \; \hat{z}_2 \sim r(\tilde{z})}[\log D(\hat{z}_1,\hat{z}_2)] \\
    &+ \mathbb{E}_{\hat{z}_1 \sim q(\tilde{z}_i|x), \; \hat{z}_2 \sim q(\tilde{z}_j|x)}[\log D(\hat{z}_1,\hat{z}_2)]\\
    &+ \mathbb{E}_{\hat{z}_1 \sim r(\tilde{z}), \; \hat{z}_2\sim q(\tilde{z}_i|x)}[\log (1-D(\hat{z}_1,\hat{z}_2))]
       \numberthis \label{eqn:diversitydis}
\end{align*}
\noindent The first term encourages the discriminator to not distinguish between samples from the prior. The second term aims to maximize overlap between different encoders, as an adversarial objective to what the generator is aiming to do in Eqn~\ref{eqn:diversitygen}. The third term minimizes overlap between the prior and each encoder.

It is important to note that the generators do not explicitly appear in the loss function because they are implicitly represented through the samples $\hat{z}_1 \sim q(\tilde{z}_i|x), \; \hat{z}_2 \sim q(\tilde{z}_j|x)$. In each SGD step we backpropagate only through the generator corresponding to the respective $(\hat{z}_1,\hat{z}_2)$ sample. We also note that we consider the pairs $(\hat{z}_1,\hat{z}_2)$ to be unordered in the losses above, because we provide both orderings to the discriminator, to ensure symmetry.   

\subsection{Overall Optimization}

The previous sub-section described the diversity inducing adversarial loss. In addition to this, we have the likelihood, and information bottleneck loss terms, denoted together by $ \mathbb{L}(\theta)$ below. Here, $\theta=(\theta_D,\theta_G,\Theta)$ denotes the parameters of the discriminator, the generators, and the decoders.

 \begin{align*}
    \mathbb{L}(\theta) &= \sum_{i=1}^m \alpha_i\mathcal{I}(\tilde{Z}_i,Y_i;\theta) -  \sum_{i=1}^m\beta_i\mathcal{I}(\tilde{Z}_i,X;\theta)  
\end{align*}



\noindent For notational convenience, we omit $\theta$ in subsequent discussions. The first term can be lower bounded, as in~\cite{vib}:
\begin{align*}
    & \mathcal{I}(\tilde{Z}_i,Y_i) \geq \int p(y_i,\tilde{z}_i)\log \frac{q(y_i|\tilde{z}_i)}{p(y_i)} \; dy_id\tilde{z}_i \numberthis \\
     &= \int  p(x)p(y_i|x)p(\tilde{z}_i|x)\log q(y_i|\tilde{z}_i) \; dx dy_i d\tilde{z}_i + H(Y) \label{eqn:firstterm}
\end{align*}
The inequality here is a result of  $KL(p(y_i|\tilde{z}_i) \; || \; q(y_i|\tilde{z}_i)) \geq 0$, where $q(y_i|\tilde{z}_i)$ is a variational approximation to the true distribution $p(y_i|\tilde{z}_i)$ and denotes our $i^{\text{th}}$ decoder. Since the entropy of output labels $H(Y)$ is independent of $\theta$, it can be ignored in the subsequent discussions. Formally, the second term can be formulated as
\begin{align*}
      \mathcal{I}(\tilde{Z}_i,X) \leq \int p(\tilde{z}_i|x)p(x) \log \frac{p(\tilde{z}_i|x)}{\psi(\tilde{z}_i)} \; d\tilde{z}_i dx \numberthis 
\end{align*}
The inequality here also results from the non-negativity of the KL divergence. The marginal $p(\tilde{z}_j)$ has been approximated by a variational approximation $\psi(\tilde{z}_j)$. Following the approach in VIB~\cite{vib}, to approximate $p(x,y_i)$ in practice we can use the empirical data-distribution $p(x,y_i)=\frac{1}{N}\sum_{n=1}^{N}\delta_{x^n}(x)\delta_{y^n_i}(y_i)$. We also note that $z^n=f(x^n)$ is the shared encoder latents, where $n$ denotes the $n^{th}$ datapoint among a total of $N$ datapoints. Now, using the re-parameterization trick, we write $\tilde{z}_i = g_i(z,\epsilon)$, where $\epsilon$ is a zero mean unit variance Gaussian noise, such that $p(\tilde{z}_i|z)=\mathcal{N}(\tilde{z}_i|g_i^\mu(z),g_i^\Sigma(z))$. We finally obtain the following lower-bound approximation of the the loss function. The detailed derivation is in the Appendix.


\begin{align*}
    \mathbb{L} &\approx \frac{1}{N}\sum_{n=1}^N \Big[\mathbb{E}_{\epsilon\sim p(\epsilon)} \Big[\sum_{i=1}^m\alpha_i \log q(y_i^n \; | \; g_i(f(x^n),\epsilon)) \\
    &- \sum_{i=1}^m\beta_i KL\Big(p(\tilde{z}_i|x^n) \; || \;  \psi(\tilde{z}_i)\Big) \Big] \Big] \numberthis \label{eqn:firsttwempirical1}
\end{align*}

\noindent In our experiments we set $\psi(\tilde{z}_j)=\mathcal{N}(\tilde{z}_j|0,\mathbb{I})$. To make predictions in classification tasks, we output the modal class of the set of class predictions by each ensemble member.

Similar to GANs~\cite{gan}, the model is optimized using alternating optimization where we alternate among objectives $\max_\theta \mathbb{L}(\theta)$, $\min_{\theta_G}L_G$, and $\max_{\theta_D}L_D$.
It is important to note that we do not explicitly optimize the KL-divergence term above, but implicitly do it during the process of adversarial learning using $\mathcal{L}_{adv}$. In Section 3.1, the case $\hat{z}_1\sim q(\tilde{z}_i|x)$ and $\hat{z}_2\sim r(\tilde{z})$ corresponds to minimizing this KL-divergence term. This is done similarly to \cite{adversarialvae}.

\subsection{Predictive uncertainty estimation}
\label{sec:aleatoric}

Our proposed method is able to meanigfully capture both epistemic and aleatoric uncertainty. Aleatoric uncertainty is typically modeled as the variance of the output distribution, which can be obtained by outputting a distribution, say a normal $p(y|x,\theta)\sim\mathcal{N}(\mu_\theta(x),\sigma_\theta(x))$~\cite{ncp}. 

Epistemic uncertainty in traditional Bayesian Neural Networks (BNNs) is captured by defining a prior (often an uninformative prior) over model weights $p(\theta)$, updating it based on the data likelihood $p(\mathcal{D}|\theta)$, where $\mathcal{D}$ is the dataset and $\theta$ is the parameters, in order to learn a posterior over model weights  $p(\theta|\mathcal{D})$. In practice, for DNNs since the true posterior cannot be computed exactly, we need to resort to samples from some approximate posterior distribution $q(\theta|\mathcal{D})\approx p(\theta|\mathcal{D})$~\cite{scalable}. 

In our approach, for epistemic uncertainty, we note that although ensembles do not consider priors over weights (unlike BNNs), they correspond to learning multiple models $\{\theta_k\}_{k=1}^K$ which can be considered to be samples from some approximate posterior $q(\theta|D)$~\cite{scalable}, where $D$ is the training dataset. We note that 
$p(y|x, \mathcal{D}) = \int_{\tilde{z}, \theta}p(y|\tilde{z}, \theta) p (\tilde{z}|x, \theta) p (\theta|\mathcal{D})$ and a typical Bayesian NN would directly approximate $p(\theta|\mathcal{D})$, which would require a prior over weights $p(\theta)$, whose selection is problematic. DIBS avoids this issue by turning sampling into optimization of a set of $\theta_k$ such that $p (\tilde{z}|x, \theta_k)$ are diverse, but still predictive of $p (\tilde{z}|x)$, without explicitly approximating $p(\theta|\mathcal{D})$. As a result there is also no notion of a \textit{true} posterior over weights $p(\theta|\mathcal{D})$ (unlike in BNNs).

For aleatoric noise, we note that we have stochastic latent variables $\tilde{z}_k\sim p(\tilde{z}_k|x)$ and obtain respective outputs $p(y_k | \tilde{z}_k)$. By sampling multiple times (say $M$ times) from $p(\tilde{z}_k|x)$, we obtain an empirical distribution $\{y_{k,i}\}_{i=1}^M$. The empirical variance of the output distributions of all the ensembles $\{y_{k,i}\}_{i=1,k=1}^{M,K}$ gives us a measure of aleatoric uncertainty 

The posterior predictive distribution gives a measure of the combined predictive uncertainty (epistemic+aleatoric), which for our approach can be calculated as follows:
\begin{align*}
    \hat{p}(y^*|x^*) = \frac{1}{MK}\sum_{k=1}^K\sum_{i=1}^M p(y^*_{k,i}|x,\theta_k)
\end{align*}
Since we enforce diversity among the $K$ ensemble members through the adversarial loss described in Section~\ref{sec:diversityloss}, we expect to obtain more reliable aleatoric uncertainty estimate and hence better predictive uncertainty overall. We perform experimental evaluation of predictive uncertainty estimation through OOD detection experiments in the next section.

\section{Experiments}

\begin{figure*}[t]
\centering
      \hspace*{-1.5em} 
        \begin{subfigure}[b]{0.35\textwidth}
               \centering
        \includegraphics[width=\textwidth]{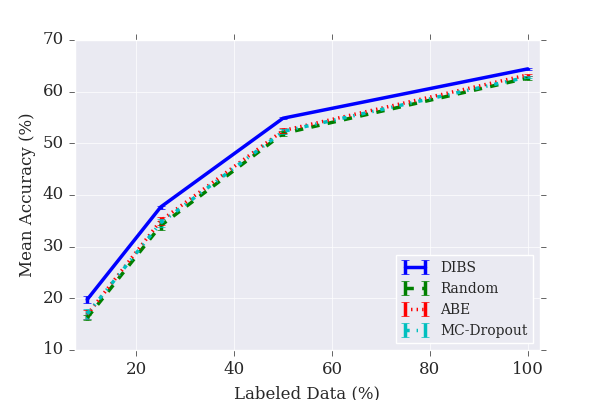}
    \caption{VGG CIFAR100}
    \label{fig:avg_r_lt}
        \end{subfigure}\hspace*{-1.3em}%
           \begin{subfigure}[b]{0.35\textwidth}
               \centering
        \includegraphics[width=\textwidth]{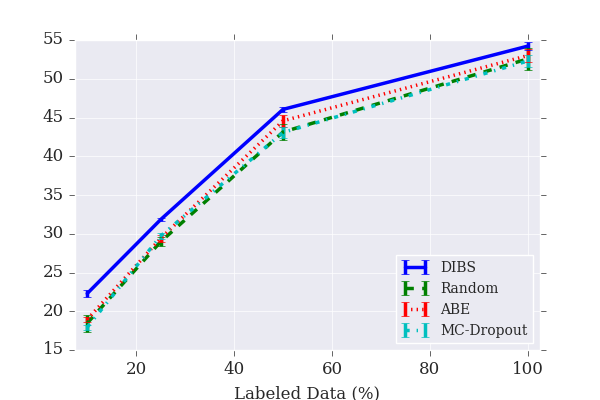}
    \caption{ResNet CIFAR100}
    \label{fig:avg_r_lt}
        \end{subfigure}\hspace*{-1.3em}%
        
           \begin{subfigure}[b]{0.35\textwidth}
               \centering
        \includegraphics[width=\textwidth]{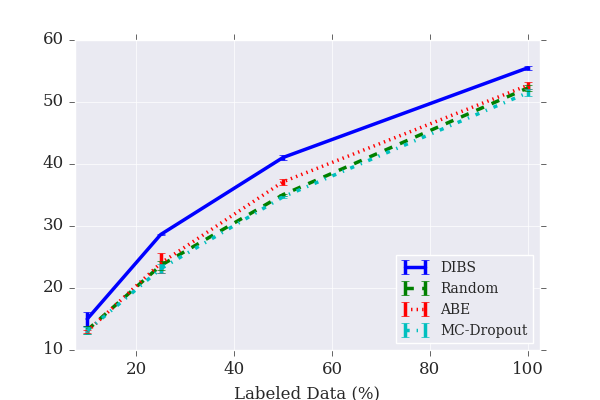}
    \caption{VGG TinyImageNet}
    \label{fig:avg_r_lt}
        \end{subfigure}\hspace*{-1.3em}%
    \begin{subfigure}[b]{0.35\textwidth}
               \centering
        \includegraphics[width=\textwidth]{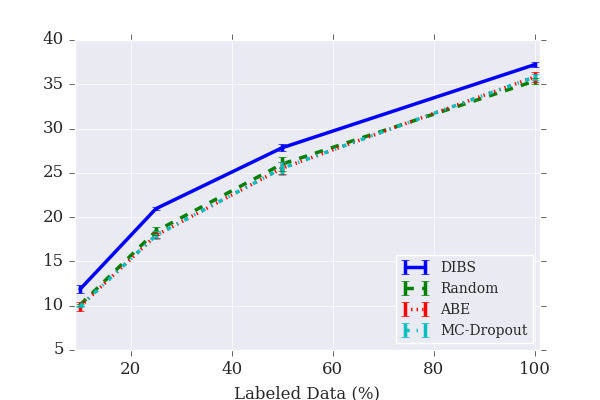}
    \caption{ResNet TinyImageNet}
    \label{fig:avg_r_lt}
        \end{subfigure}%
         \begin{subfigure}[b]{0.35\textwidth}
               \centering
        \includegraphics[width=\textwidth]{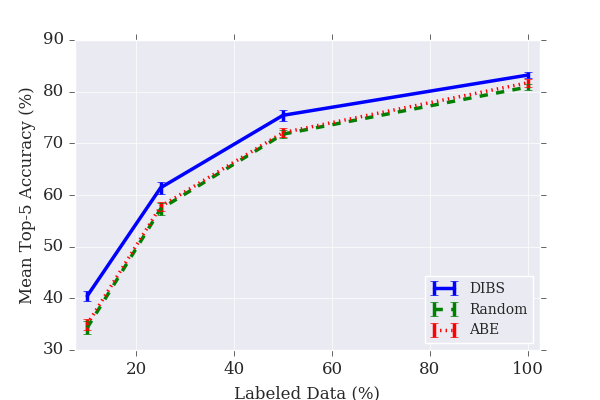}
    \caption{VGG MITPlaces2}
    \label{fig:avg_r_lt}
        \end{subfigure}%
    \quad
     \caption{Performance of baselines, Random~\cite{lakshminarayanan2017simple}, MC-Dropout~\cite{mcdropout} and ABE~\cite{deepmetriclearning} against our proposed approach \algoName on four datasets with two backbone architectures. All results show $\%$ accuracy on the test dataset. The y-axis label on (a) propagates to figures (b), (c), (d), and (e). We show results for different $\%$ of labels of the dataset used during training. It is evident that when less data is used for training, \algoName relatively performs much better than the baseline. The specific details of the architectural variants are in the Appendix.}   \label{fig:generalization}%
\end{figure*}

In the section, we show how our method is able to achieve:
\begin{enumerate}[
    topsep=0pt,
    noitemsep,
    leftmargin=*,
    itemindent=12pt]
\item \textbf{Better accuracy}:  How do the proposed approach and baselines perform on the task of image classification in the face of limited data?
\item \textbf{Better generalization}: How well do the models generalize when the evaluation distribution is different from the training distribution?
\item \textbf{Better uncertainty estimation}: Are we are able to obtain better uncertainty estimates compared to the baselines as evidenced by OOD detection?
\end{enumerate}
We compare our approach to four external baselines: \textbf{ABE}~\cite{deepmetriclearning}, \textbf{NCP}~\cite{ncp}, \textbf{MC-Dropout}~\cite{mcdropout}, and the state-of-the-art deep ensemble scheme of~\cite{lakshminarayanan2017simple}, that considers ensembles to be randomly initialized and trained neural networks. We henceforth call this method \textbf{Random}. For NCP, we impose the NCP priors on the input and output for each NN architecture that we evaluate. For images, the input prior amounts to an additive Gaussian noise on each pixel. ABE~\cite{deepmetriclearning} considers a diversity inducing loss based on pairwise squared difference among the ensemble outputs, and is a recently proposed strong baseline. 
MC-Dropout~\cite{mcdropout} is a Bayesian method that samples dropout masks repeatedly to produce different predictions from the model.
We evaluate the performance of these baselines along our model \algoName in five benchmark datasets: MNIST~\cite{mnist}, CIFAR-10, CIFAR-100~\cite{cifar},TinyImageNet~\cite{tiny,imagenet}, and MIT Places 2~\cite{places}. 
Performing experiments on the Places 2 dataset confirms that our method is able to scale well to large scale settings as it is a scene recognition dataset with over 1.8 M images and 365 unique classes.

\subsection{Experimental Setup}
\label{sec:setup}
We run experiments with two standard vision architectures as the backbone, namely VGG19~\cite{vgg} and ResNet18~\cite{resnet}. 
For optimization, we use Stochastic Gradient Descent (SGD)~\cite{sgd} with a learning rate of 0.05 and momentum of 0.9~\cite{momentum}. 
We decay the learning rate by a factor of 10 every 30 epochs of training. 

\subsection{Performance}
\noindent \textbf{Experiments on CIFAR10, CIFAR100, TinyImageNet, and MIT Places 2 show that \algoName outperforms all baselines on the task of image classification}. We evaluate the performance of the proposed approach \algoName and all the baselines on four datasets: MNIST,  CIFAR100, TinyImageNet, and MIT Places 2 and evaluating the classification accuracy. To demonstrate good performance ``at-scale," we consider three base architectures: a simple 4-layer CNN, VGG Networks~\cite{vgg}, and ResNets~\cite{resnet}. Specifics of these architectures are mentioned in the Appendix. Fig.~\ref{fig:generalization} show results in terms of $\%$ accuracy on the CIFAR-10, CIFAR-100, TinyImageNet, and MIT Places 2 datasets when there are respectively $100\%,50\%,25\%$, and $10\%$ of the labeled dataset used during training. For the MIT Places 2 dataset, we considered the top-5 classification accuracy in order to be consistent with the evaluation metric in the original challenge~\cite{places}. For all the other datasets, we consider the top-1 classification accuracy. We randomly sampled examples from the entire training dataset to create these smaller training sets. 

It is interesting to note that when less data is used during training, \algoName performs relatively much better than the baselines indicating better generalization. As evident from Fig.~\ref{fig:generalization}, \algoName consistently performs better than all the baseline schemes with all the base architectures. The results on the Places 2 dataset demonstrates that our approach can effectively scale to a significantly larger dataset. From Fig.~\ref{fig:generalization}, we can also see that the relevant magnitude of performance improvement of \algoName over baselines increase as the dataset size increases (MIT Places 2, TinyImageNet, CIFAR100). This suggests the efficacy of our approach in the image classification task.

\subsection{Generalization and Transfer experiments}
\label{generalization}
In this section we consider experiments of generalization to changes in the data distribution (without finetuning) and transfer under dataset shift to a different test distribution (with finetuning). For all the experiments here we use a simple 4 layer feedforward CNN with maxpool layer and ReLU non-linearity after every layer. Details are mentioned in the Appendix. We use this instead of a VGGNet or ResNet due to the small scale of the datasets involved in the experiments.
\vspace*{-0.1cm}
\subsubsection{Generalization to in-distribution changes:}
\noindent \textbf{\algoName effectively generalizes to dataset change under translation, and rotation of digits}. In this section, we consider the problem of generalization, through image translation, and image rotation experiments on MNIST. The  generalization experiments on MNIST are described below:
\begin{enumerate}[
    topsep=0pt,
    noitemsep,
    leftmargin=*,
    itemindent=12pt]
\item \textbf{Translate (Trans)}: Training on normal MNIST images, and testing by randomly translating the images by 0-5 pixels, 0-8 pixels, and 0-10 pixels respectively. 
\item \textbf{Rotate}: Training on normal MNIST digits, and testing by randomly rotating the images by 0-30 degrees and 0-45 degrees respectively. 
\item \textbf{Interpolation-Extrapolation-Translate (IETrans)}: We train on images translated randomly in the range [-5,5] pixels and test on images translated randomly in the range [-10,10] pixels.
\item \textbf{Interpolation-Extrapolation-Rotate (IERotate)}: We train train on images rotated randomly in the range [-22,22] degrees and test on images rotated randomly in the range [-45,45] degrees.
\item \textbf{Color}: We train on Normal MNIST images and test on colored MNIST images~\cite{cmnist} by randomly changing foreground color to red, green, or blue.
\end{enumerate}

The interpolation-extrapolation experiments help us understand the generalization of the models on the data distribution that it was trained on (interpolation), as well as on a data distribution objectively different from training (extrapolation). Hence, we consider the testing distributions to be a superset of the training distribution in these two experiments. Table~\ref{tb:mnist} summarizes the results of these experiments.  We observe that \algoName achieves over $2\%$ higher accuracy compared to the baseline of Random~\cite{lakshminarayanan2017simple} in the experiment of generalizing under translation shift and over $1\%$ higher accuracy in the rotational shift experiment.
\vspace*{-0.1cm}
\subsubsection{Transfer under dataset shift}

\noindent \textbf{Here we show that \algoName effectively transfers when trained on a source dataset and finetuned and evaluated on a target dataset}. We train our model \algoName on one dataset which we call the source and finetune it on another dataset which we call the target, and finally evaluate it on the test set of the target dataset. We consider the following experiment: \begin{enumerate}[
    topsep=0pt,
    noitemsep,
    leftmargin=*,
    itemindent=12pt]
\item \textbf{Source}: MNIST~\cite{mnist}; \textbf{Target}: SVHN~\cite{svhn} 
\end{enumerate}

For finetuning in the target dataset, we fix the encoder(s) of \algoName and the baselines Random~\cite{lakshminarayanan2017simple} and ABE~\cite{deepmetriclearning}, and update the parameters of the decoders  for a few fixed iterations. We train on MNIST for 50 epochs and fine-tune on the training datset of SVHN for 20 epochs before evaluating on the test dataset of SVHN. Details of the exact procedure are in the Appendix. Results in Table~\ref{tb:mnist} show that \algoName achieves higher accuracy under transfer to the target environment in this experiment as compared to the baselines.

\setlength{\tabcolsep}{2pt}
\begin{table}[t]
\centering
\begin{tabular}{@{}ccccc@{}}
\toprule
 \multicolumn{5}{c}{\textbf{Generalization experiments}} \\
 \midrule
\textbf{Exp}                 & \textbf{Test details} & \textbf{Random}& \textbf{ABE} & \textbf{\algoName} \\ \midrule
\multirow{3}{*}{\textbf{Trans}} & \textbf{[-5,5] p}      & 76.38\footnotesize{$\pm$0.70}    &   77.53\footnotesize{$\pm$0.68}    & \textbf{79.46\footnotesize{$\pm$0.50}}           \\
                                    & \textbf{[-10,10] p}     & 37.84\footnotesize{$\pm$1.05}     &   39.06\footnotesize{$\pm$0.98}    & \textbf{41.38\footnotesize{$\pm$1.00}}           \\ \midrule
\multirow{2}{*}{\textbf{Rotate}}    & \textbf{[-30,30] d}    & 95.79\footnotesize{$\pm$0.18}     &  94.96\footnotesize{$\pm$0.18}     & \textbf{96.36\footnotesize{$\pm$0.11}}           \\
                                    & \textbf{[-45,45] d}    & 87.09\footnotesize{$\pm$0.31}     &  87.82\footnotesize{$\pm$0.16}     & \textbf{88.90\footnotesize{$\pm$0.17}}    \\ \midrule
                                \multirow{1}{*}{\textbf{IETrans}} & \textbf{[-10,10] p}      &  93.04\footnotesize{$\pm$0.22}     &  92.76\footnotesize{$\pm$0.30}    &      \textbf{94.21\footnotesize{$\pm$0.41}}   \\                                \midrule
                                 
\multirow{1}{*}{\textbf{IERotate}}    & \textbf{[-45,45] d}    & 97.79\footnotesize{$\pm$0.10}      &  97.85\footnotesize{$\pm$0.15}    & \textbf{98.35\footnotesize{$\pm$0.37}}          \\ \midrule
\textbf{Color} & R,G,B & 97.63\footnotesize{$\pm$0.49}      &  97.61\footnotesize{$\pm$0.38}    & \textbf{98.20\footnotesize{$\pm$0.63}}\\  \midrule
\multicolumn{5}{c}{\textbf{Transfer experiments}}\\\midrule
\textbf{Source}                 & \textbf{Target} & \textbf{Random}& \textbf{ABE} & \textbf{\algoName} \\\midrule
\multirow{1}{*}{\textbf{MNIST}}    & \textbf{SVHN}    &   43.11\footnotesize{$\pm$2.10}   &47.05\footnotesize{$\pm$1.62}    & \textbf{50.09\footnotesize{$\pm$0.97}}\\   \bottomrule      
\end{tabular}
\caption{Generalization and Transfer experiments on MNIST. Details of the experiments are mentioned in Section~\ref{generalization}. Results show that for all the experiments, \algoName outperforms the baselines. For the transfer experiment, we train for 50 epochs on MNIST, free the encoder(s) and fine-tune in the training dataset of SVHN for 20 epochs and report the $\%$ accuracy on the test dataset of SVHN. Here p denotes pixel and d degrees. The backbone is a simple 4-layer CNN described in the Appendix.}
\label{tb:mnist}
\end{table}

\setlength{\tabcolsep}{5pt}
\begin{table*}[t]
\centering
\begin{minipage}[c]{0.7\textwidth}
\begin{tabular}{@{}cccccc@{}}
\toprule
 
\multicolumn{6}{c}{\textbf{Trained on CIFAR-10/ Trained on CIFAR-100}}                                                                             \\ \midrule                                                                                                                                                                                                                                  

\multirow{6}{*}{\textbf{AUROC}} & \textbf{}        & \textbf{\begin{tabular}[c]{@{}c@{}}ImageNet\\ Crop\end{tabular}} & \textbf{\begin{tabular}[c]{@{}c@{}}ImageNet\\ Resize\end{tabular}} & \textbf{Gaussian} & \textbf{Uniform}  \\ \midrule
&\textbf{Random} & 84.61/68.87                                                       & 80.53/73.14                                                         & 77.09/68.24             & 77.06/68.24               \\

&\textbf{NCP}  &              83.13/66.13                                          &  78.11/68.93                                                       &       75.31/72.43      & 75.21/72.10                \\
&\textbf{ABE}  &     85.24/\textbf{69.94}				                                                 &   	79.09/73.45                                                      &    	82.16/76.45       &  82.24/76.32          \\
&\textbf{\algoName}  & \textbf{87.57}/\textbf{70.06}                                                        & \textbf{82.47}/\textbf{75.71}                                                          & \textbf{86.56}/\textbf{80.61}             & \textbf{86.56}/\textbf{80.13}                   \\ 
\bottomrule
\multirow{6}{*}{\textbf{AUPR}} &  \textbf{}        & \textbf{\begin{tabular}[c]{@{}c@{}}ImageNet\\ Crop\end{tabular}} & \textbf{\begin{tabular}[c]{@{}c@{}}ImageNet\\ Resize\end{tabular}} & \textbf{Gaussian} & \textbf{Uniform} \\ \midrule
&\textbf{Random}           & 82.08/67.72                                                        & 77.76/73.69                                                          & 88.52/84.66             & 88.52/84.66            \\

&\textbf{NCP}           &                  82.60/66.15                                     &  75.34/70.16                                                        &      85.29/84.06       & 85.16/83.49         \\
&\textbf{ABE}       &         	84.33/\textbf{68.11}				                                            &       77.71/\textbf{74.86}                                          &   89.10/89.21        &   89.94/89.08      \\
&\textbf{\algoName}              & \textbf{88.67}/\textbf{68.28}                                                        & \textbf{81.10}/\textbf{74.98}                                                          & \textbf{93.36}/\textbf{91.00}             & \textbf{93.30}/\textbf{91.52}            \\ 
\bottomrule
\end{tabular}
\end{minipage}
\begin{minipage}[c]{0.28\textwidth}
\caption{Comparison of performance on OOD data classification. \algoName and the baselines, Random, ABE, and NCP
are trained on a particular dataset (CIFAR-10/CIFAR-100) and are then tasked with prediction of images to be in-distribution or out-of-distribution (OOD). For OOD examples, we use four datasets as described in Section~\ref{sec:ood}. For evaluation, we use the metrics AUROC (Area Under the Receiver Operating Characteristic Curve) and AUPR (Area under the Precision-Recall curve). Results show that \algoName significantly outperforms all the baselines on both the metrics.} 
\label{tb:ood}
\end{minipage}
\end{table*}

\vspace*{-0.2cm}
\subsection{Uncertainty estimation through OOD examples}
\label{sec:ood}
\vspace*{-0.05cm}

\noindent \textbf{\algoName achieves accurate predictive uncertainty estimates necessary for reliable OOD detection}.
We follow the scheme of~\cite{ncp,ood} for evaluating on Out-of-Distribution (OOD) examples. We train our model on CIFAR~\cite{cifar} and then at test time, consider images sampled from the dataset to be in-distribution and images sampled from a different dataset, say Tiny-Imagenet to be OOD. In our experiments, we use four OOD datasets, namely Imagenet-cropped~\cite{imagenet} (randomly cropping image patches of size 32x32), Imagenet-resized~\cite{imagenet} (downsampling images to size 32x32), synthetic uniform-noise, and Gaussian-Noise. The details are same as~\cite{ood}.


To elaborate on the specifics of OOD detection at test time, given input image $\mathbf{x}$, we calculate the softmax score of the input with respect to each ensembles $\mathcal{S}_i(\mathbf{x})$ and compare the score to a threshold $\delta_i$. For \algoName, each $\mathcal{S}_i(\mathbf{x})$ corresponds to a particular decoder head. 
The aggregated ensemble prediction is given by the mode of the individual predictions. The details of this procedure are mentioned in the Appendix.
Table~\ref{tb:ood} compares the performance of \algoName against baselines, Random~\cite{lakshminarayanan2017simple} and NCP~\cite{ncp}. It is evident that \algoName consistently outperforms both the baselines on both the AUROC and AUPR metrics.

\vspace{-0.2cm}
\section{Related Work}
\label{sec:related}
Ensembles have been used in fields ranging from computer vision~\cite{snapshot,multiplechoicedhruv} to reinforcement learning and imitation learning for planning and control~\cite{pets,infogail}. Traditionally, ensembles have been proposed to tackle the problem of effective generalization~\cite{ensemblesold1}, and algorithms like random forests~\cite{forest}, and broad-approaches like boosting~\cite{boosting}, and bagging~\cite{bagging} are common ensemble learning techniques. In ensemble learning, multiple models are trained to solve the same problem.  Each individual learner model is a simple model, or a `weak learner' while the aggregate model is a `strong learner.'

Diversity among ensemble learners, important for generalization~\cite{multiplechoicedhruv,ensemblesold2,ensemblesold1}, has traditionally been ensured by training each weak learner on a separate held-out portion of the training data (bagging)~\cite{bagging}, adding random noise to the output predictions, randomly initializing model weights~\cite{lakshminarayanan2017simple,anonymouslosslandscape}, having stochastic model weights~\cite{bnn}, or by manipulating the features and attributes~\cite{deepmetriclearning,multiplechoicedhruv} of the model. As demonstrated by~\cite{lakshminarayanan2017simple}, bagging is not a good diversity inducing mechanism, when the underlying base learner has multiple local optima, as is the case with neural net architectures, which are the focus of this paper. BNNs~\cite{bnn} provide reasonable epistemic uncertainty estimates but do not necessarily capture the inherent aleatoric uncertainty, and so are not capable of successfully inducing diversity in the output predictions for effectively modeling multi-modal data~\cite{anonymouslosslandscape,aleatoricvision}.

~\cite{lakshminarayanan2017simple} proposes a mechanism of randomly initializing the weights of a neural net architecture, and hence obtaining an ensemble of neural network models, treated as an uniformly weighted mixture of Gaussians. This approach outperforms bagging and BNNs in terms of both predictive accuracy and uncertainty estimation, however, as pointed out in~\cite{anonymouslosslandscape} the number of ensembles needed to accurately identify different modes and model each mode sufficiently requires a large number of models, and is computationally expensive. 

Motivated by this, instead of adopting a random initialization approach, we proposed a principled scheme of diversity maximization among latent ensemble variables, so that different modes in the data distribution are identified, and constrained the diversity of the latent variables through an information bottleneck.
We adopted the approach of having a shared encoder and $K-$ headed stochastic decoder, with each head of the decoder representing one model of the ensemble and utilize an adversarial loss to promote meaningful diversity. 
\cite{deepmetriclearning} proposes a similar architecture, but for enforcing diversity among the decoders, the authors explicitly maximize the Euclidean distance between every pair of feature embeddings (for each datapoint), and is not guaranteed to separate the multiple data modes ``in-distribution" in the embedding space.  



Another important component of our architecture is an information bottleneck constraint, that constrains the flow of information from the input layer $X$ to each of the $K$ stochastic latent decoder variables $\tilde{Z}_i$'s, so that the predictions don't become arbitrarily diverse due to the diversity inducing loss. This relates to the work in~\cite{vib}, which we extend to $K$ latent variables instead of just one.


\vspace*{-0.2cm}
\vspace*{-0.2cm}
\section{Conclusion}
\vspace*{-0.2cm}
In this paper we addressed the issue of enforcing diversity in a learned ensemble through a novel adversarial loss, while ensuring high likelihood of the predictions, through the notion of variational information bottleneck. We demonstrate through extensive experimentation that the proposed approach outperforms state-of-the-art baseline ensemble and Bayesian learning methods on four benchmark datasets in terms of accuracy under sparse training data, uncertainty estimation for OOD detection, and generalization to a test distribution significantly different from the training data distribution. Our technique is generic and applicable to any latent variable model.

\clearpage 
{\small
\bibliography{references}
}
\clearpage
\section{Appendix}
\subsection{OOD detection details}

\noindent \textbf{DIBS achieves accurate predictive uncertainty estimates necessary for reliable OOD detection}.
We follow the scheme of~\cite{ncp,ood} for evaluating on Out-of-Distribution (OOD) examples. We train our model on a particular dataset, say CIFAR-10~\cite{cifar} and then at test time, consider images sampled from CIFAR-10 to be in-distribution and images sampled from a different dataset, say Mini-Imagenet to be OOD. In our experiments, we use four OOD datasets, namely Imagenet-cropped~\cite{imagenet} (randomly cropping image patches of size 32x32), Imagenet-resized~\cite{imagenet} (downsampling images to size 32x32), a synthetic uniform-noise dataset, and a synthetic Gaussian-Noise dataset. The details of these are same as in~\cite{ood}.

In our experiments, we use four OOD datasets, namely Imagenet-cropped~\cite{imagenet} (randomly cropping image patches of size 32x32), Imagenet-resized~\cite{imagenet} (downsampling images to size 32x32), a synthetic uniform-noise dataset, and a synthetic Gaussian-Noise dataset. In the uniform-noise dataset, there are 10000 images with each pixel sampled from a unifrom distribution on [0,1]. In the Gaussian-noise dataset, there are 10000 io ages with each pixel sampled from an i.i.d. Gaussian with 0.5 mean and unit variance. All pixels are clipped to be in the range [0,1]. For evaluation, we use the metrics AUROC (Area Under the Receiver Operating Characteristic Curve)~\cite{auroc} and AUPR (Area under the Precision-Recall curve)~\cite{aupr1,aupr2}.

To elaborate on the specifics of OOD detection at test time, given input image $\mathbf{x}$, we calculate the softmax score of the input with respect to each ensembles $\mathcal{S}_i(\mathbf{x})$ and compare the score to a threshold $\delta_i$. For DIBS, each $\mathcal{S}_i(\mathbf{x})$ corresponds to a particular decoder head. So, the individual OOD detectors are given by:
\begin{equation}
     Q_i(\mathbf{x};\delta_i)=
    \begin{cases}
      1, & \text{if}\ \mathcal{S}_i(\mathbf{x})\leq \delta_i \\
      0, & \text{otherwise}
    \end{cases}
  \end{equation}
Here, $1$ denotes an OOD example. The aggregated ensemble prediction is given by the mode of the individual predictions:
\begin{equation}
     \mathbf{Q}(\mathbf{x};\delta_i)=
    \begin{cases}
      1, & \text{if}\ \sum_{i=1}^kQ_i(\mathbf{x};\delta_i) \geq0.5\\
      0, & \text{otherwise}
    \end{cases}
  \end{equation}
Since all the ensembles are ``equivalent," so we set all $\delta_i=\delta$ for the experiments. We choose the same $\delta$ values as reported in Figure 13 of the ODIN paper~\cite{ood}. We can also apply the temperature scaling and input pre-processing heuristics in ODIN~\cite{ood} to DIBS and the baselines so as to potentially obtain better OOD detection. However, we do not do this for our experiments so as to unambiguously demonstrate the benefit of the ensemble approach alone. Table 2 in the paper compares the performance of DIBS against baselines, Random~\cite{lakshminarayanan2017simple} and NCP~\cite{ncp}. It is evident that DIBS consistently outperforms both the baselines on both the AUROC and AUPR metrics.

\subsection{GANs, Adversarial Autoencoders}
\label{sec:gan}
Adversarial Autoencoders (AAEs) use GANs~\cite{gan} for structuring the latent space of an autoencoder such that the encoder learns to convert the data-distribution to the prior distribution and the decoder learns to map the prior to the data distribution. Instead of constraining the latent space $\tilde{Z}$ to be close to the prior $p(Z)$ through a KL-divergence as done in VAEs~\cite{vae}, this paper describes that training a discriminator through adversarial loss helps in fitting better to the multiple modes of the data distribution. \Florian{Maybe worth mentioning implicit generative models, of which AAE is a special case}\HB{GANs in general are implicit generative models right?https://arxiv.org/pdf/1610.03483.pdf}. Inspired by this paper, we develop a novel diversity-inducing objective, that enforces the stochastic latent variables of each ensemble member to be different from each other through a discriminator trained through an adversarial objective.

In a GAN, a generator $G(z)$ is trained to map samples $z$ from a prior distribution $p(z)$ to the data distribution $\hat{p}(x)$, while ensuring that the generated samples maximally confuse a discriminator $D(x)$ into thinking they are from the true data distribution $p(x)$. The optimization objective can be summarized as:
\begin{align*}
    \min_G\max_D \mathbb{E}_{x\sim p(x)}[\log D(x)] + \mathbb{E}_{z\sim p(z)}[\log(1-D(G(z))] 
\end{align*}
In AAEs,  for the discriminator $D$, the true (real) data samples come from a prior $p(z)$, while the generated (fake) samples come from the posterior latent state distribution $q_\theta(\tilde{z};\theta)$, where $q_\theta(\tilde{z};\theta)=\int q(\tilde{z}|x;\theta)p(x) dx$. In Section 3.1 we describe our diversity inducing loss which is inspired by this formulation.

\subsection{Overall Objective}
The previous sub-section described the diversity inducing adversarial loss. In addition to this, we have the likelihood, and information bottleneck loss terms, denoted together by $ \mathbb{L}(\theta)$ below. Here, $\theta=(\theta_D,\theta_G,\Theta)$ denotes the parameters of the discriminator, the generators, and the decoders.
 \begin{align*}
    \mathbb{L}(\theta) &= \sum_{i=1}^m \alpha_i\mathcal{I}(\tilde{Z}_i,Y_i;\theta) -  \sum_{i=1}^m\beta \mathcal{I}(\tilde{Z}_i,X;\theta)  
\end{align*}



\noindent For notational convenience, we omit $\theta$ in subsequent discussions. The first term can be lower bounded, as in~\cite{vib}:
\begin{align*}
     \mathcal{I}(\tilde{Z}_i,Y_i) &\geq \int p(y_i,\tilde{z}_i)\log \frac{q(y_i|\tilde{z}_i)}{p(y_i)} \; dy_id\tilde{z}_i \numberthis \\
     &= \int  p(x)p(y_i|x)p(\tilde{z}_i|x)\log q(y_i|\tilde{z}_i) \; dx dy_i d\tilde{z}_i + H(Y) 
\end{align*}
The inequality here is a result of  $KL(p(y_i|\tilde{z}_i) \; || \; q(y_i|\tilde{z}_i)) \geq 0$, where $q(y_i|\tilde{z}_i)$ is a variational approximation to the true distribution $p(y_i|\tilde{z}_i)$ and denotes our $i^{\text{th}}$ decoder. Since the entropy of output labels $H(Y)$ is independent of $\theta$, it can be ignored in the subsequent discussions. Formally, the second term can be formulated as
\begin{align*}
      \mathcal{I}(\tilde{Z}_i,X) \leq \int p(\tilde{z}_i|x)p(x) \log \frac{p(\tilde{z}_i|x)}{\psi(\tilde{z}_i)} \; d\tilde{z}_i dx \end{align*}
The inequality here also results from the non-negativity of the KL divergence. The marginal $p(\tilde{z}_j)$ has been approximated by a variational approximation $\psi(\tilde{z}_j)$. Following the approach in VIB~\cite{vib}, to approximate $p(x,y_i)$ in practice we can use the empirical data-distribution $p(x,y_i)=\frac{1}{N}\sum_{n=1}^{N}\delta_{x^n}(x)\delta_{y^n_i}(y_i)$. We also note that $z^n=f(x^n)$ is the shared encoder latents, where $n$ denotes the $n^{th}$ datapoint among a total of $N$ datapoints. The first two terms of the overall loss $ \mathbb{L}(\theta)$ are $\leq$ the following variational bound
\begin{align*}
    \mathbb{L}_1(\theta) &= \sum_{i=1}^m\alpha_i \int dx dy_i d\tilde{z}_i p(y_i|x)p(\tilde{z}_i|x)\log q(y_i|\tilde{z}_i)   \\ 
    &- \sum_{j=1}^m\beta_i\int  d\tilde{z}_j dz p(\tilde{z}_j|x)p(x) \log \frac{p(\tilde{z}_j|x)}{\psi(\tilde{z}_j)} 
\end{align*}

Now, using the re-parametrization trick, we write $\tilde{z}_i = g_i(z,\epsilon)$, where $\epsilon$ is a zero mean unit variance Gaussian noise, such that $p(\tilde{z}_i|z)=\mathcal{N}(\tilde{z}_i|g_i^\mu(z),g_i^\Sigma(z))$.

\begin{align*}
    \mathbb{L}_1(\theta) &\approx \frac{1}{N}\sum_{n=1}^N\left[\sum_{i=1}^m\alpha_i \int d\tilde{z}_i p(\tilde{z}_i|x^n)\log q(y_i^n|\tilde{z}_i) \right] \\ 
    &- \frac{1}{N}\sum_{n=1}^N\left[ \sum_{j=1}^m\beta_i\int  d\tilde{z}_j p(\tilde{z}_j|x^n) \log \frac{p(\tilde{z}_j|x^n)}{\psi(\tilde{z}_j)}  \right] 
\end{align*}
We finally obtain the following lower-bound approximation of the the loss function.
\begin{align*}
    \mathbb{L}(\theta) &\approx \frac{1}{N}\sum_{n=1}^N[\mathbb{E}_{\epsilon\sim p(\epsilon)}[\sum_{i=1}^m\alpha_i \log q(y_i^n|g_i(f(x^n),\epsilon)) \\
    &- \sum_{i=1}^m\beta_i KL[p(\tilde{Z}_i|x^n),\psi(\tilde{Z}_i)]] + \mathcal{L}_{adv}
\end{align*}


\noindent In our experiments we set $\psi(\tilde{z}_j)=\mathcal{N}(\tilde{z}_j|0,\mathbb{I})$. To make predictions in classification tasks, we output the modal class of the set of class predictions by each ensemble member. For regression tasks, we output the average prediction in the ensemble.

It is important to note that we do not explicitly optimize the KL-divergence term above, but implicitly do it during the process of adversarial learning using $\mathcal{L}_{adv}$. In Section 3.1, the case $\hat{z}_1\sim q(\tilde{z}_i|x)$ and $\hat{z}_2\sim r(\tilde{z})$ corresponds to minimizing this KL-divergence term. This is inspired by the AAE paper that we described in the previous section of this Appendix.

\subsection{Training details}
The small neural network used for the experiments in Table 1 consists of 4 convolutional layers and ReLU non-linearities~\cite{cnn}.  The discriminator used for adversarial training of the proposed diversity loss is a 4 layer MLP (Multi-Layered Perceptron). For optimization we use ADAM with a learning rate of 0.0001. For the hyperparameters $\alpha_i$ and $\beta_i$, we set all $\beta_i=\beta$ and all $\alpha_i= 1-\beta$ and perform gridsearch for $\beta$ in the range $[10^{-4},10^{-1}]$. We found $\beta=10^{-2}$ to work the best and the results reported in the paper are with this value. The code will be released soon and a link posted on the first authors' websites.

\begin{figure}

 \caption{(a) Plot showing \algoName consistently outperforming baselines on the test TinyImageNet dataset by varying the number of ensemble heads $K$ during training.}
\includegraphics[width=6cm]{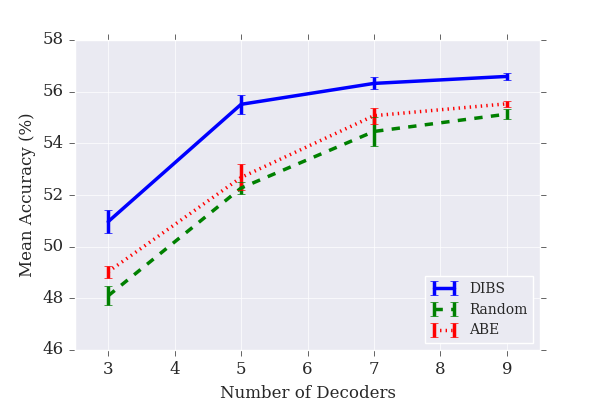}\label{fig:hyper1}
\end{figure}

\subsection{Experiments with \algoName variations}
\label{sec:misc}

\noindent \textbf{Experiments showing \algoName is efficient to train, and trains high likelihood ensembles}. In this section, we perform some experiments to understand \algoName better. We compare the performance of \algoName by varying $K$ i.e. the number of decoder heads, which translates to the number of model ensembles. We show that by varying $K$, there isn't a significant performance gain after a certain threshold value of $K$, say $K^*$. In  Figure~\ref{fig:hyper1}, $K^*$ is around 8, and it is interesting to note that \algoName consistently outperforms the baselines for all values of $K$.

\end{document}